\documentclass[conference]{IEEEtran}
\IEEEoverridecommandlockouts
\usepackage[nocompress]{cite}
\AtBeginDocument{}
\usepackage[utf8]{inputenc}
\usepackage[T1]{fontenc}
\usepackage{amsmath,amssymb,amsfonts}
\usepackage{graphicx}
\usepackage{textcomp}
\usepackage{booktabs}
\usepackage{url}
\usepackage{microtype}
\usepackage[caption=false,font=footnotesize]{subfig}
\usepackage{float}
\usepackage[hidelinks]{hyperref}

\begin{document}

\title{On the Effectiveness and Limitations of Open-Weight LLMs for Low-Resource Offensive Language Detection in Danish}

\author{
\IEEEauthorblockN{Elnur Adl Zarabi}
\IEEEauthorblockA{Texas A\&M University-Corpus Christi\\
\texttt{eadlzarabi@islander.tamucc.edu}}
\and
\IEEEauthorblockN{Amit Kumar}
\IEEEauthorblockA{Texas A\&M University-Corpus Christi\\
\texttt{akumar3@islander.tamucc.edu}}
\and
\IEEEauthorblockN{Adi Singh}
\IEEEauthorblockA{Mississippi State University\\
\texttt{as5142@msstate.edu}}
\and
\IEEEauthorblockN{Sierra Nelson}
\IEEEauthorblockA{Mississippi State University\\
\texttt{Sierra.nelson@ssrc.msstate.edu}}
\and
\IEEEauthorblockN{Megan Stubbs-Richardson}
\IEEEauthorblockA{University of Miami\\
\texttt{Mxr3845@miami.edu}}
\and
\IEEEauthorblockN{Taoran Ji}
\IEEEauthorblockA{Texas A\&M University-Corpus Christi \\
\texttt{taoran.ji@tamucc.edu}}
}

\maketitle

\begin{abstract}
We study the effectiveness and robustness of open-weight large language models (LLMs) for Danish offensive language detection, an underexplored low-resource task.
We evaluate four model families (Llama, Mistral, Gemma, and Qwen) across eleven prompting strategies spanning five design categories, cross-prompt generalization, Danish orthographic variation, and robustness under balanced and naturally imbalanced class distributions.
We compare against three BERT-family baselines with distinct pre-training strategies, evaluated via both masked language model prompting and supervised fine-tuning.
Even without task-specific fine-tuning, LLMs can surpass all three fine-tuned BERT baselines, and lightweight finetuning yields substantial further gains.
No single model or prompt category dominates, underscoring that prompt design interacts with model pre-training in ways that cannot be predicted a priori.
Semantically grounded prompts transfer more robustly, while format-constrained patterns consistently fail to generalize.
ASCII transliteration of Danish-specific characters (\texttt{æ}, \texttt{ø}, \texttt{å}) generally degrades performance, indicating these characters encode meaningful signal.
At the same time, the gap between a balanced test set (best F1 of 0.84) and the corpus's natural class skew (0.62 trained on skewed data, 0.70 trained on balanced data) -- relative drops of 26.2\% and 16.7\% -- shows that class imbalance remains a critical open challenge.
Our code and configurations are available at \url{https://github.com/PASHAADC/Danish-LLM-HateSpeech}.
\end{abstract}

\section{Introduction}

\begin{figure*}[!t]
    \centering
    \includegraphics[width=\textwidth]{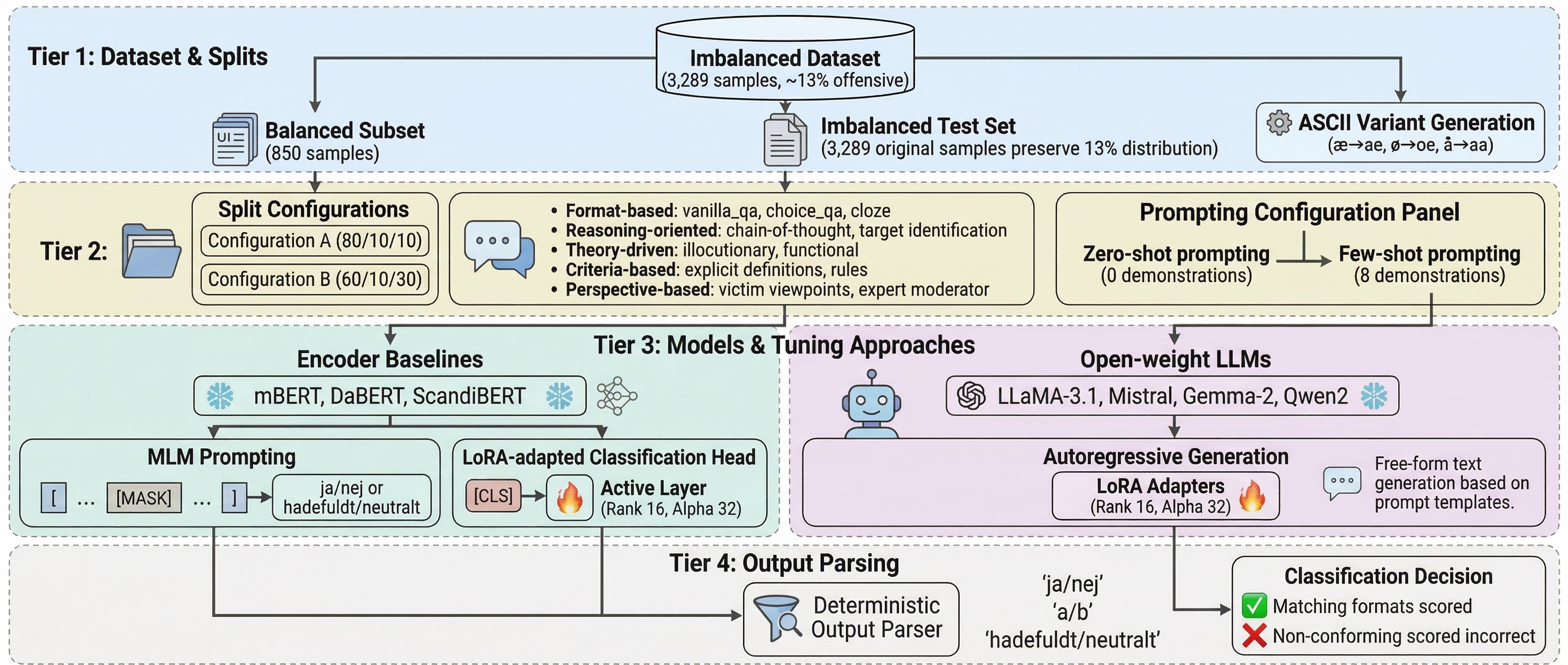}
\caption{Experimental pipeline for evaluating the effectiveness and robustness of Danish offensive language detection models. From the DKHate corpus, we derive balanced and naturally imbalanced settings, construct ASCII-transliterated variants for orthographic robustness analysis, and apply two stratified split configurations (80/10/10 and 60/10/30). These evaluation settings are used to compare BERT-family baselines and LoRA-adapted open-weight LLMs across multiple prompting regimes, with final outputs mapped to task labels through deterministic parsing.}
    \label{fig:pipeline}
\end{figure*}

The proliferation of online hate speech poses urgent societal challenges demanding effective automated detection.
Danish represents a critical yet understudied case as a low-resource language with approximately six million speakers and lacks the annotated corpora and specialized models available for English \cite{kirkedal-etal-2019-lacunae}.
Danish statistics document increased hate crimes with approximately half occurring on digital platforms \cite{rigspolitiet2024hate}, concerning given causal links between social media activity and real-world hate crimes \cite{mueller-schwarz-2021-fanning}.
Open-weight LLMs offer a promising path forward due to their customizability, transparency, and sensitivity.
However, their effectiveness for Danish offensive language detection remains unexamined.
To the best of our knowledge, this paper presents the first systematic investigation of LLM efficacy for Danish offensive language detection. Figure~\ref{fig:pipeline} provides an overview of the experimental pipeline, including dataset construction, prompting-based evaluation of BERT-family baselines and open-weight LLMs, and robustness analysis under balanced and naturally imbalanced settings.

Research on LLM-based hate speech detection has focused almost exclusively on English, revealing that effectiveness depends on many factors. Zhang et al. \cite{zhang-etal-2024-dont-go} show that LLMs exhibit excessive sensitivity to group-related linguistic cues, frequently misclassifying benign statements as hateful, while also displaying poor calibration with confidence scores concentrated in narrow ranges regardless of task difficulty. Melis et al. \cite{melis2025taxonomy} organize hate speech definitions into a taxonomy of 14 conceptual elements and demonstrate that varying the specificity of the definition provided in the prompt substantially affects zero-shot classification performance, though the effect is inconsistent across model architectures. Roy et al. \cite{roy-etal-2023-probing} vary the content of zero-shot prompts for GPT-3.5 and FLAN-T5 on three English benchmarks and find that supplying the targeted community lifts F1 by 20 to 30 percent over a plain query while definitions help one model and hurt another, so the value of a prompt ingredient depends on the model that reads it. Plaza-del-Arco et al. \cite{plaza-del-arco-etal-2023-respectful} prompt encoder and instruction-tuned models zero-shot on eight English benchmarks and on Spanish and Italian, where the choice of label words alone shifts macro-F1 by several points and FLAN-T5 matches a fine-tuned RoBERTa on three datasets, though zero-shot stays below the fine-tuned baseline in both low-resource languages. Prompt design plays a critical role more broadly \cite{white2023promptpatterncatalogenhance}, with explanation-enhanced strategies proving particularly effective. Nghiem and Daumé III \cite{nghiem-daume-iii-2024-hatecot} show that pretraining on chain-of-thought explanations drawn from over 52,000 samples significantly improves generalization across offensive speech benchmarks in both zero-shot and few-shot settings. Class imbalance also remains central, as hate speech constitutes only a small fraction of real-world content. Tonneau et al. \cite{tonneau2025hatedayinsightsglobalhate} find that hateful posts represent roughly 2--3\% of a representative day on Twitter, underscoring the severity of the class skew that detection systems must handle.

Research on Danish hate speech detection remains limited.
Sigurbergsson and Derczynski \cite{sigurbergsson-derczynski-2020-offensive} introduced the DKHate corpus and benchmarked it using logistic regression, SVM, and BiLSTM classifiers.
Müller-Eberstein et al. \cite{muellereberstein2025dakultur} show that native Danish data more than doubles LLM acceptance rates versus translated alternatives, highlighting cross-lingual transfer limitations and motivating the use of language-specific resources.
On the BERT side, several models have become available---multilingual BERT \cite{devlin2019bertpretrainingdeepbidirectional}, the monolingual DaBERT \cite{danishbert2020}, and the regional ScandiBERT \cite{snaebjarnarson-etal-2023-transfer}---but their effectiveness for Danish offensive language detection remains unexamined, making them natural baselines for contextualizing LLM performance.
To address these gaps, we systematically evaluate four open-weight LLM families and three BERT baselines across eleven prompting patterns, three evaluation settings, and two split configurations. Our main findings are:
\begin{itemize}
    \item BERT models evaluated via both masked language modeling (MLM) prompting and supervised fine-tuning reveal that monolingual Danish pre-training (DaBERT) substantially outperforms multilingual and regional alternatives on imbalanced data when paired with task-specific supervision. ScandiBERT's nominally highest balanced-data score is obtained by predicting the offensive class for every input (recall of 1.0), i.e., a degenerate classifier rather than learned discrimination. MLM prompting proves fundamentally limited for this task.
    \item Under balanced evaluation, open-weight LLMs achieve strong performance with minimal fine-tuning. Imbalanced distributions reveal substantial degradation, with no single model or prompt dominating. Theory-informed prompts often remain comparatively robust under class imbalance, while format-constrained patterns consistently fail to generalize.
    \item ASCII transliteration of Danish characters (æ, ø, å $\rightarrow$ ae, oe, aa) generally degrades performance, indicating that these characters encode meaningful signal that should be preserved.
    \item Zero-shot and few-shot fine-tuning achieve comparable peak performance, but few-shot gains are less consistent, suggesting in-context examples do not reliably improve fine-tuned representations.
\end{itemize}

\section{Methodology}
\label{sec:methodology}

This section presents the framework we use to study Danish offensive
language detection. We construct datasets under balanced, naturally
imbalanced, and cross-balanced settings, we benchmark BERT-family models
and open-weight large language models on those datasets, we compare
prompting strategies and fine-tuning regimes, and we probe robustness to
orthographic variation and to social media noise, while class imbalance
enters through the settings themselves. Four research questions organize
the work. How do open-weight large language models compare to
BERT-family baselines under balanced and naturally imbalanced settings?
How well do fine-tuned models generalize across prompting patterns?
How does ASCII transliteration of Danish-specific characters affect
performance? Do explicit denoising instructions improve robustness to
social media noise? Figure~\ref{fig:pipeline} shows the full workflow.

\subsection{Dataset}\label{subsec:dataset}
We use the DKHate corpus \cite{sigurbergsson-derczynski-2020-offensive}, an established Danish offensive language benchmark comprising 3,289 samples collected from Facebook and Reddit, annotated with binary labels: \textit{offensive} and \textit{not offensive}. We use \textit{hadefuld tale} as a natural Danish rendering of the task question; while Danish usage also includes terms such as \textit{stødende sprog} (offensive language), our findings concern relative patterns across models and prompts rather than absolute sensitivity to any single lexical choice.
To systematically assess model behavior under conditions reflecting different deployment scenarios, we define three evaluation settings.
Each setting constitutes an independent fine-tuning and evaluation pipeline, partitioned into training, validation, and test sets via stratified sampling under two split ratios: $\text{config}_{\text{a}}$ (80/10/10) and $\text{config}_{\text{b}}$ (60/10/30).
Under $\text{config}_{\text{a}}$ the balanced and imbalanced test sets hold 85 and 329 samples, and under $\text{config}_{\text{b}}$ 255 and 987, with the full split sizes and the split files themselves in the repository.

The three settings differ in how class balance is distributed.
1)~\textit{Balanced}: all 425 offensive instances are paired with 425 randomly sampled non-offensive instances, yielding 850 samples with a 50/50 class ratio.
This setting isolates model and prompt effects from class-distribution confounds.
2)~\textit{Imbalanced}: the full 3,289-sample corpus is used with its natural distribution of approximately 13\% offensive content. This share sits well above the 2--3\% of hateful posts that Tonneau et al. \cite{tonneau2025hatedayinsightsglobalhate} measure on a representative day of Twitter, for two reasons that lie in the corpus rather than in Danish. The DKHate guidelines follow the OLID scheme and label any post containing profanity as offensive, so the positive class is broader than hate speech, and 232 of the original 3,600 comments were retrieved through a lexicon of offensive terms rather than sampled at random \cite{sigurbergsson-derczynski-2020-offensive}. The imbalanced setting is therefore the milder of the two skews, and a detector on an unfiltered stream would face a rarer positive class than any we evaluate.
3)~\textit{Balanced-Train / Imbalanced-Test}: models are trained and validated on the balanced subset but evaluated on the imbalanced test set.
This setting mirrors a realistic deployment scenario in which a model is trained on a curated, balanced corpus and must generalize to the naturally skewed distribution at test time.
Crucially, the imbalanced test set is drawn exclusively from samples not included in the balanced training or validation splits.

Additionally, for each setting, we construct an ASCII-transliterated variant by deterministically replacing \texttt{æ}, \texttt{ø}, and \texttt{å} with \texttt{ae}, \texttt{oe}, and \texttt{aa}, respectively.
This allows us to isolate the effect of orthographic normalization on model performance (Section~\ref{subsec:rq3}).

\subsection{Models}
We evaluate one representative model from each of four prominent open-weight LLM families: LLaMA-3.1-8B-Instruct \cite{grattafiori2024llama3herdmodels}, Mistral-7B-Instruct-v0.3 \cite{jiang2023mistral7b}, Gemma-2-9B-it \cite{gemma2024}, and Qwen2-7B-Instruct \cite{qwen2024}.
All four models are multilingual in practice: in our preliminary experiments, each exhibits at least partial Danish capability and can generate Danish text, though output fluency and consistency vary, suggesting non-trivial exposure to Danish during pre-training without large-scale supervised fine-tuning or reinforcement learning from human feedback specifically targeting Danish, let alone Danish offensive language detection.
This observation is consistent with the respective technical reports.
These properties make the selected models suitable test cases for studying how general-purpose open-weight LLMs transfer to Danish offensive language detection in the absence of the extensive fine-tuning and alignment typically available for high-resource languages such as English.
Each LLM is fine-tuned via prompt-completion-style supervised fine-tuning.
Training examples are formatted as single-turn conversations in which a Danish-language prompt asks whether a given text constitutes hate speech, and the target completion is a single token: Ja (yes) or Nej (no). Each LLM is adapted with LoRA \cite{hu2022lora} (rank 16, $\alpha=32$, dropout 0.05) on the attention and feed-forward projections of every decoder layer, trained with AdamW at a learning rate of $2\times10^{-4}$ for up to 10 epochs with early stopping on validation loss, and decoded with nucleus sampling at temperature 1.0 and top-$p$ 0.9. The training prompt, \texttt{Er følgende tekst hadefuld tale? Svar kun med Ja eller Nej.}, is the same for every model, and the full configurations, prompt templates, and formatting examples are in the repository.

Additionally, to contextualize LLM performance, we include three BERT-family baselines, each representing a distinct pre-training strategy commonly employed for low-resource language tasks:
1) multilingual BERT (mBERT) \cite{devlin2019bertpretrainingdeepbidirectional}, a 110M-parameter model pre-trained on 104 languages, representing broad multilingual coverage;
2) DaBERT \cite{danishbert2020}, a 110M-parameter model pre-trained on 9.5GB of Danish text with a dedicated 32K Danish BPE vocabulary, representing monolingual Danish pre-training;
   and 3) ScandiBERT \cite{snaebjarnarson-etal-2023-transfer}, a 124M-parameter XLM-RoBERTa model pre-trained on five Scandinavian languages, representing regional multilingual pre-training on closely related languages.
   Each BERT model is evaluated under two fine-tuning strategies: (i) CLS: standard sequence classification with a linear head over the [CLS] token; and (ii) MLM: a masked language modeling formulation that mirrors the LLM prompting setup, where the model predicts Ja or Nej at a masked answer position.
   Both use the LoRA setup of the LLMs on the query, key, value, and dense projections, the CLS head is trained with class-weighted cross-entropy at a learning rate of $2\times10^{-5}$ with the checkpoint of highest validation F1 kept, and the input formats for both formulations are in the repository. Runs whose CLS head collapsed to a near-constant prediction were retrained from scratch up to five times and the attempt with the highest test F1 was retained, so the CLS figures in Table~\ref{tab:bert-cls} are selected on the test split and are optimistic for that reason.

\subsection{Evaluation Design}
We evaluate the following eleven prompting patterns, grouped into five categories, each designed to test whether a specific cognitive framing elicits different classification behavior.
These patterns are applied to all LLMs and MLM-fine-tuned BERT models.
CLS-fine-tuned BERT models are instead evaluated v ia standard classification, where the classification head directly predicts the label without any prompting. For the LLMs, each pattern also has an 8-shot form that prepends eight fixed hand-written exemplars, four offensive and four not and none of them drawn from the corpus, to the instruction.

The Danish templates for all eleven patterns, with English translations, are in the repository.
\textbf{Format-based} patterns vary output structure: \textit{vanilla QA} expects a direct ja/nej answer, \textit{choice QA} presents labeled options (A/B), and \textit{cloze} asks the model to complete a sentence with the appropriate label \cite{zhang-etal-2024-dont-go}. These impose minimal interpretive burden and serve as controlled baselines, testing whether accuracy depends on output format rather than task understanding.

\textbf{Reasoning-oriented} patterns require explanations alongside classification: \textit{chain-of-thought} asks for a verdict followed by an explanation -- an answer-first variant of chain-of-thought prompting \cite{wei2023chainofthoughtpromptingelicitsreasoning}; because the verdict precedes the reasoning, it does not test reasoning-first deliberation -- and \textit{target identification} asks which group is affected. These test whether explicit deliberation improves detection of subtle or context-dependent instances.

\textbf{Theory-driven} patterns ground classification in linguistic frameworks: \textit{illocutionary} prompts ask what speech act the text performs \cite{austin1962how}, and \textit{functional} prompts ask whether the text excludes others or establishes social dominance. These test whether theoretical anchoring provides more robust criteria than format-driven instructions, particularly under class imbalance where surface cues become unreliable.

\textbf{Criteria-based} patterns supply an explicit operational definition of offensive language with a checklist of conditions drawn from the DKHate annotation guidelines \cite{sigurbergsson-derczynski-2020-offensive}, testing whether models can follow structured rubrics rather than relying on learned associations.

\textbf{Perspective-based} patterns shift the evaluative stance: \textit{victim perspective} instructs the model to judge from the target's point of view \cite{chiril2022emotionally}, and \textit{expert moderator} assigns a content-moderation persona. These test whether adopting a role grounded in empathy or domain expertise changes classification behavior.

For MLM-fine-tuned BERT models, the answer position is masked and the predicted label is taken as the higher-probability token between Ja and Nej.

For all prompt-based evaluations, a deterministic parser matches model outputs against the expected format (e.g., ja/nej for vanilla QA, A/B for choice QA). Non-conforming outputs are mapped to the negative (not-offensive) class rather than excluded; they therefore count against the model on offensive gold labels, while on non-offensive gold labels they are indistinguishable from correct rejections. Table~\ref{tab:parse-failure} reports their prevalence from a re-parse of the saved zero-shot outputs, with no re-training or re-inference, counting as non-conforming every output without a parseable verdict in the expected format, such as an empty output, a prompt echo, an unfilled mask, degenerate token repetition, or a refusal. The cloze pattern is largely unparseable for all four models (86--98\% non-conforming), as models rarely complete the masked template with a clean label, while the remaining format failures concentrate in specific model--pattern combinations rather than being uniform, with Gemma on the illocutionary (96\%) and vanilla QA (41\%) patterns as the clearest cases. Pooled across patterns, 21.5\% of the outputs that the pipeline scored as the negative class were in fact non-conforming, so a share of the near-zero F1 scores below reflects format non-compliance rather than misclassification.

\begin{table}[!t]
\caption{Non-conforming output rate (\%) per model and pattern, zero-shot, pooled over the balanced and imbalanced evaluations. The parser scores these as the negative class.}
\label{tab:parse-failure}
\centering
\small
\setlength{\tabcolsep}{4pt}
\renewcommand{\arraystretch}{1.05}
\begin{tabular}{@{}lrrrr@{}}
\toprule
\textbf{Pattern} & \textbf{LLaMA} & \textbf{Mistral} & \textbf{Gemma} & \textbf{Qwen} \\
\midrule
vanilla QA          & 3  & 2  & 41 & 13 \\
choice QA           & 1  & 2  & 26 & 1  \\
cloze               & 86 & 98 & 88 & 91 \\
cot                 & 7  & 2  & 4  & 3  \\
target              & 4  & 2  & 5  & 14 \\
illocutionary       & 5  & 3  & 96 & 8  \\
functional          & 2  & 2  & 3  & 9  \\
definition          & 6  & 5  & 2  & 3  \\
victim perspective  & 2  & 2  & 6  & 6  \\
expert moderator    & 3  & 2  & 9  & 5  \\
rules               & 1  & 2  & 8  & 22 \\
\bottomrule
\end{tabular}
\end{table}

\section{Results}
We conduct a total of 1,466 experiments spanning four LLMs, three BERT baselines, eleven prompting patterns, three evaluation settings (Balanced, Imbalanced, Balanced-Train / Imbalanced-Test), and two split configurations ($\text{config}_a$ and $\text{config}_b$).
Unless stated otherwise, P, R, and F1 denote precision, recall, and F1 of the \emph{offensive} (positive) class. Table~\ref{tab:corrected-metrics} additionally reports macro-averaged F1 recomputed on conforming outputs only, as a check on whether the parser artifacts of Table~\ref{tab:parse-failure} change the picture. The correction is at most 0.02 except for base Gemma under balanced evaluation ($+0.07$) and fine-tuned Qwen ($+0.05$ balanced, $+0.04$ imbalanced), so the ordering of the base models is unchanged, while the two Qwen corrections lift fine-tuned Qwen above Gemma under balanced evaluation and above Mistral under imbalanced evaluation without touching the per-pattern positive-class results on which our conclusions rest. The coverage column shows that fine-tuned LLaMA (26\%) and fine-tuned Qwen (49\%) reach their balanced-evaluation scores on a minority of parseable outputs, which we flag as a caveat on those two configurations.

\begin{table}[!t]
\caption{Reported vs.\ corrected macro-F1 (mean over all patterns) and format coverage. Corrected F1 uses conforming outputs only.}
\label{tab:corrected-metrics}
\centering
\small
\setlength{\tabcolsep}{4pt}
\renewcommand{\arraystretch}{1.05}
\begin{tabular}{@{}llrrrr@{}}
\toprule
\textbf{Setting} & \textbf{Model} & \textbf{Reported} & \textbf{Corrected} & \textbf{$\Delta$} & \textbf{Coverage} \\
\midrule
\multicolumn{6}{l}{\textit{Balanced, base}} \\
& LLaMA   & .53 & .55 & $+.02$ & .89 \\
& Mistral & .52 & .54 & $+.02$ & .89 \\
& Gemma   & .59 & .66 & $+.07$ & .75 \\
& Qwen    & .52 & .52 & $+.01$ & .84 \\
\multicolumn{6}{l}{\textit{Balanced, fine-tuned}} \\
& LLaMA   & .62 & .61 & $-.01$ & .26 \\
& Mistral & .48 & .48 & $-.00$ & .84 \\
& Gemma   & .65 & .65 & $+.00$ & .90 \\
& Qwen    & .63 & .68 & $+.05$ & .49 \\
\multicolumn{6}{l}{\textit{Imbalanced, base}} \\
& LLaMA   & .46 & .45 & $-.01$ & .89 \\
& Mistral & .47 & .46 & $-.01$ & .89 \\
& Gemma   & .57 & .56 & $-.01$ & .74 \\
& Qwen    & .58 & .57 & $-.01$ & .84 \\
\multicolumn{6}{l}{\textit{Imbalanced, fine-tuned}} \\
& LLaMA   & .51 & .51 & $+.00$ & .71 \\
& Mistral & .45 & .45 & $-.01$ & .88 \\
& Gemma   & .57 & .57 & $+.01$ & .90 \\
& Qwen    & .43 & .47 & $+.04$ & .71 \\
\bottomrule
\end{tabular}
\end{table}

\begin{table}[!t]
\caption{CLS-fine-tuned BERT results; best F1 per condition in bold. Rows with recall 1.0 are degenerate all-positive classifiers (F1 $\approx 0.66$ by construction on a 50/50 split). MLM results are in the repository.}
\label{tab:bert-cls}
\centering
\small
\setlength{\tabcolsep}{2pt}
\renewcommand{\arraystretch}{1.05}
\begin{tabular}{@{}ll ccc @{\hspace{6pt}} ccc @{}}
\toprule
\multicolumn{2}{c}{}
& \multicolumn{3}{c}{\textbf{Balanced}}
& \multicolumn{3}{c}{\textbf{Imbalanced}} \\
\cmidrule(lr){3-5} \cmidrule(lr){6-8}
\textbf{Model} & \textbf{Configuration}
& \textbf{P} & \textbf{R} & \textbf{F1}
& \textbf{P} & \textbf{R} & \textbf{F1} \\
\midrule
mBERT      & $\text{config}_a$ (80/10/10)  & .51 & .67 & .58 & .22 & .49 & .31 \\
DaBERT     & $\text{config}_a$ (80/10/10)  & .51 & .76 & .61 & .29 & .60 & \textbf{.39} \\
ScandiBERT & $\text{config}_a$ (80/10/10)   & .49 & 1.0 & \textbf{.66} & .13 & 1.0 & .23 \\
\midrule
mBERT      & $\text{config}_b$ (60/10/30)  & .60 & .51 & .55 & .27 & .38 & .31 \\
DaBERT     & $\text{config}_b$ (60/10/30)   & .47 & .48 & .48 & .26 & .55 & \textbf{.35} \\
ScandiBERT & $\text{config}_b$ (60/10/30)   & .50 & 1.0 & \textbf{.66} & .15 & .94 & .25 \\
\bottomrule
\end{tabular}
\end{table}

\subsection{Primary Results}

\begin{table*}[!t]
\caption{Zero-shot performance without fine-tuning, $\text{config}_a$ (80/10/10). Best F1 per setting in bold.}
\label{tab:zeroshot-base-A}
\centering
\small
\setlength{\tabcolsep}{2pt}
\renewcommand{\arraystretch}{1.05}
\begin{tabular}{@{}l*{12}{c}@{\hspace{4pt}}*{12}{c}@{}}
\toprule
& \multicolumn{12}{c}{\textbf{Balanced}}
& \multicolumn{12}{c}{\textbf{Imbalanced}} \\
\cmidrule(lr){2-13} \cmidrule(lr){14-25}
& \multicolumn{3}{c}{\textbf{LLaMA}} & \multicolumn{3}{c}{\textbf{Mistral}} & \multicolumn{3}{c}{\textbf{Gemma}} & \multicolumn{3}{c}{\textbf{Qwen}} & \multicolumn{3}{c}{\textbf{LLaMA}} & \multicolumn{3}{c}{\textbf{Mistral}} & \multicolumn{3}{c}{\textbf{Gemma}} & \multicolumn{3}{c}{\textbf{Qwen}} \\
\cmidrule(lr){2-4} \cmidrule(lr){5-7} \cmidrule(lr){8-10} \cmidrule(lr){11-13} \cmidrule(lr){14-16} \cmidrule(lr){17-19} \cmidrule(lr){20-22} \cmidrule(lr){23-25}
\textbf{Pattern} & \textbf{P} & \textbf{R} & \textbf{F1} & \textbf{P} & \textbf{R} & \textbf{F1} & \textbf{P} & \textbf{R} & \textbf{F1} & \textbf{P} & \textbf{R} & \textbf{F1} & \textbf{P} & \textbf{R} & \textbf{F1} & \textbf{P} & \textbf{R} & \textbf{F1} & \textbf{P} & \textbf{R} & \textbf{F1} & \textbf{P} & \textbf{R} & \textbf{F1} \\
\midrule
vanilla      & .46 & .45 & .46 & .54 & .52 & \textbf{.53} & .71 & .40 & .52 & .58 & .26 & .36 & .15 & .65 & .25 & .19 & .63 & .29 & .30 & .35 & .32 & .35 & .44 & \textbf{.39} \\
choice       & .54 & .88 & \textbf{.67} & .45 & .81 & .58 & .61 & .55 & .57 & .61 & .55 & .57 & .12 & .81 & .21 & .13 & .86 & .22 & .23 & .53 & .32 & .25 & .60 & \textbf{.35} \\
cloze        & .62 & .36 & .45 & .50 & .02 & .05 & .79 & .36 & .49 & .64 & .43 & \textbf{.51} & .17 & .33 & .22 & .18 & .09 & .12 & .24 & .30 & \textbf{.27} & .16 & .33 & .22 \\
cot          & .50 & .36 & .42 & .53 & .71 & .61 & .66 & .60 & \textbf{.62} & .50 & .12 & .19 & .12 & .37 & .18 & .16 & .74 & .26 & .22 & .60 & \textbf{.33} & .19 & .19 & .19 \\
target       & .50 & .60 & .54 & .50 & .67 & .57 & .61 & .67 & \textbf{.64} & .66 & .45 & .54 & .15 & .65 & .24 & .13 & .58 & .21 & .23 & .79 & \textbf{.35} & .23 & .47 & .31 \\
illocution.  & .60 & .62 & .61 & .64 & .67 & \textbf{.65} & .86 & .14 & .24 & .64 & .55 & .59 & .22 & .60 & .32 & .26 & .70 & \textbf{.38} & .36 & .12 & .18 & .25 & .65 & .37 \\
functional   & .47 & .33 & .39 & .67 & .69 & \textbf{.68} & .64 & .64 & .64 & .51 & .55 & .53 & .28 & .63 & .39 & .24 & .72 & .36 & .27 & .74 & \textbf{.40} & .25 & .74 & .38 \\
definition   & .71 & .52 & .60 & .49 & .88 & \textbf{.63} & .60 & .64 & .62 & .29 & .05 & .08 & .11 & .21 & .14 & .14 & .91 & .25 & .23 & .72 & \textbf{.35} & .12 & .07 & .09 \\
victim       & .56 & .52 & .54 & .57 & .48 & .52 & .63 & .64 & \textbf{.64} & .70 & .38 & .49 & .17 & .60 & .26 & .22 & .56 & .32 & .27 & .70 & .39 & .38 & .49 & \textbf{.42} \\
expert       & .67 & .52 & .59 & .67 & .69 & \textbf{.68} & .71 & .40 & .52 & .59 & .31 & .41 & .14 & .28 & .19 & .31 & .77 & .44 & .22 & .47 & .29 & .38 & .56 & \textbf{.45} \\
rules        & .17 & .02 & .04 & .83 & .24 & \textbf{.37} & .88 & .17 & .28 & .43 & .07 & .12 & .03 & .02 & .03 & .22 & .26 & \textbf{.23} & .28 & .19 & .22 & .16 & .07 & .10 \\
\bottomrule
\end{tabular}
\end{table*}

This subsection reports the primary results for Danish offensive language detection.
Fine-tuned LLM results are reported in Tables~\ref{tab:zeroshot-configA}, \ref{tab:zeroshot-configB}, \ref{tab:cross-balanced}, and~\ref{tab:fewshot-finetuned}.
CLS-fine-tuned BERT results are reported in Table~\ref{tab:bert-cls}, while the per-pattern MLM-fine-tuned BERT results are in the repository.
Zero-shot performance without task-specific fine-tuning under $\text{config}_a$ is reported in Table~\ref{tab:zeroshot-base-A}, and the corresponding $\text{config}_b$ table and the 8-shot tables without fine-tuning are in the repository, from which we quote best values below.

First, we observe that general-purpose open-weight LLMs already surpass fine-tuned BERT baselines on Danish offensive language detection without any task-specific fine-tuning, provided a suitable prompting pattern is used.
Specifically, under balanced evaluation, multiple zero-shot LLM–pattern combinations (Table~\ref{tab:zeroshot-base-A} for $\text{config}_a$) exceed the best CLS-fine-tuned BERT result (Table~\ref{tab:bert-cls}; ScandiBERT, 0.66 F1 in both configurations -- a score this baseline obtains degenerately, by predicting the offensive class for every input): Mistral reaches 0.68 F1 in $\text{config}_a$ and 0.74 F1 in $\text{config}_b$. Against the best \emph{non-degenerate} fine-tuned BERT results (DaBERT, 0.61 F1 in $\text{config}_a$; mBERT, 0.55 F1 in $\text{config}_b$), these constitute relative improvements of 11.5\% and 34.5\%, respectively; differences of the smaller magnitudes involved here should nonetheless be interpreted with care given the test-set sizes (see Limitations).
Under imbalanced evaluation with skewed class distribution, the gap persists.
The best LLM–pattern combination achieves 0.45 F1 (Qwen, Table~\ref{tab:zeroshot-base-A}) and 0.42 F1 (Gemma, $\text{config}_b$), while the best CLS-fine-tuned BERT achieves only 0.39 F1 and 0.35 F1, respectively --- relative improvements of 15.4\% and 20.0\%.
Furthermore, a similar observation holds when comparing with the MLM-prompting BERT baselines, whose best F1 reaches 0.68 under balanced evaluation and 0.24 under imbalanced evaluation, and whose balanced-evaluation scores again come from near all-positive predictions.
Additionally, these results further corroborate our earlier observation that these models were exposed to Danish during pre-training and possess a non-trivial degree of understanding of Danish language structure.

Second, we observe that the lightweight fine-tuning strategy described in Section~\ref{sec:methodology} yields substantial improvements in LLM performance.
Under balanced evaluation in $\text{config}_a$, the best fine-tuned LLM result rises to 0.84 F1 (Gemma, Table~\ref{tab:zeroshot-configA}) from 0.68 F1 without fine-tuning (Mistral, Table~\ref{tab:zeroshot-base-A}), a relative improvement of 23.5\%.
Under imbalanced evaluation in $\text{config}_a$, the improvement is even more pronounced: the best result increases from 0.45 F1 (Qwen, Table~\ref{tab:zeroshot-base-A}) to 0.62 F1 (Gemma, Table~\ref{tab:zeroshot-configA}), a relative gain of 37.8\%. Similar trends hold for $\text{config}_b$.
These gains suggest that even minimal task-specific fine-tuning on a dataset of this scale offers a promising path for low-resource offensive language detection.

Third, we find that few-shot learning gains occur more frequently in raw LLMs than in fine-tuned ones. Among fine-tuned models, 55\% of cases (Tables~\ref{tab:zeroshot-configA}, \ref{tab:zeroshot-configB} and \ref{tab:fewshot-finetuned}) show improved performance under few-shot over zero-shot evaluation. In contrast, 65\% of cases show such improvement without fine-tuning (Table~\ref{tab:zeroshot-base-A} and the $\text{config}_b$ and 8-shot tables in the repository).

We attribute this to diminishing returns: fine-tuning already adapts the model to the task, leaving less room for in-context examples to contribute. Prior to fine-tuning, the LLMs possess structural understanding of Danish but lack task-specific adaptation, allowing few-shot exemplars to provide a greater marginal benefit.

Fourth, we observe that some LLM families exhibit preferences for specific prompting categories, suggesting an interaction between model pre-training and cognitive framing style.
For instance, across Tables~\ref{tab:zeroshot-configA}, \ref{tab:zeroshot-configB}, and~\ref{tab:fewshot-finetuned}, LLaMA consistently achieves relatively strong performance on the functional prompting pattern, while Mistral's preference is volatile and prone to collapse on certain prompts (e.g., chain-of-thought in Table~\ref{tab:zeroshot-configB}).
Additionally, we find that despite using a vanilla QA-style prompt-completion fine-tuning strategy, models do not necessarily perform best when evaluated on the vanilla QA pattern.
Although counterintuitive, this is understandable: these LLMs have been extensively fine-tuned and aligned via RLHF on high-resource languages across diverse tasks, and our lightweight adaptation is unlikely to alter their ingrained behavioral preferences.
Instead, it appears to guide each model to leverage its inherent understanding of Danish, acquired during pre-training, to perform this specific task more effectively.

\begin{table*}[!t]
\caption{Zero-shot results of the fine-tuned models, $\text{config}_a$ (80/10/10). Best F1 per setting in bold.}
\label{tab:zeroshot-configA}
\centering
\small
\setlength{\tabcolsep}{2pt}
\renewcommand{\arraystretch}{1.05}
\begin{tabular}{@{}l*{12}{c}@{\hspace{4pt}}*{12}{c}@{}}
\toprule
& \multicolumn{12}{c}{\textbf{Balanced}}
& \multicolumn{12}{c}{\textbf{Imbalanced}} \\
\cmidrule(lr){2-13} \cmidrule(lr){14-25}
& \multicolumn{3}{c}{\textbf{LLaMA}} & \multicolumn{3}{c}{\textbf{Mistral}} & \multicolumn{3}{c}{\textbf{Gemma}} & \multicolumn{3}{c}{\textbf{Qwen}} & \multicolumn{3}{c}{\textbf{LLaMA}} & \multicolumn{3}{c}{\textbf{Mistral}} & \multicolumn{3}{c}{\textbf{Gemma}} & \multicolumn{3}{c}{\textbf{Qwen}} \\
\cmidrule(lr){2-4} \cmidrule(lr){5-7} \cmidrule(lr){8-10} \cmidrule(lr){11-13} \cmidrule(lr){14-16} \cmidrule(lr){17-19} \cmidrule(lr){20-22} \cmidrule(lr){23-25}
\textbf{Pattern} & \textbf{P} & \textbf{R} & \textbf{F1} & \textbf{P} & \textbf{R} & \textbf{F1} & \textbf{P} & \textbf{R} & \textbf{F1} & \textbf{P} & \textbf{R} & \textbf{F1} & \textbf{P} & \textbf{R} & \textbf{F1} & \textbf{P} & \textbf{R} & \textbf{F1} & \textbf{P} & \textbf{R} & \textbf{F1} & \textbf{P} & \textbf{R} & \textbf{F1} \\
\midrule
vanilla     & .77 & .55 & .64 & .53 & .83 & .65 & .82 & .86 & \textbf{.84} & .61 & .90 & .73 & .00 & .00 & .00 & .00 & .00 & .00 & .21 & .88 & \textbf{.33} & .14 & .84 & .24 \\
choice      & .48 & .69 & .57 & .57 & .67 & .62 & .78 & .76 & \textbf{.77} & .73 & .76 & .74 & .15 & .56 & .24 & .20 & .47 & .28 & .26 & .63 & \textbf{.37} & .16 & .93 & .27 \\
cloze       & .57 & .88 & \textbf{.69} & .72 & .31 & .43 & .72 & .62 & .67 & .49 & .88 & .63 & .15 & .40 & .22 & .00 & .00 & .00 & .58 & .35 & \textbf{.43} & .19 & .47 & .27 \\
cot         & .57 & .86 & .69 & .53 & .79 & .63 & .64 & .88 & .74 & .74 & .88 & \textbf{.80} & .00 & .00 & .00 & .21 & .21 & .21 & .13 & .67 & .22 & .36 & .63 & \textbf{.45} \\
target      & .74 & .83 & \textbf{.79} & .46 & .71 & .56 & .91 & .24 & .38 & .53 & .93 & .67 & .17 & .70 & .27 & .15 & .81 & .25 & .25 & .58 & \textbf{.35} & .14 & .09 & .11 \\
illocution.  & .52 & .76 & .62 & .49 & 1.00 & .66 & .80 & .48 & .60 & .63 & .86 & \textbf{.73} & .30 & .51 & \textbf{.38} & .14 & .63 & .22 & .83 & .12 & .20 & .17 & .88 & .29 \\
functional  & .56 & .95 & .71 & .44 & .17 & .24 & .63 & .52 & .57 & .83 & .71 & \textbf{.77} & .30 & .44 & \textbf{.36} & .15 & .77 & .25 & .69 & .21 & .32 & .16 & .98 & .27 \\
definition  & .83 & .12 & .21 & .52 & .60 & .56 & .49 & .55 & .52 & .52 & .95 & \textbf{.67} & .21 & .60 & \textbf{.31} & .16 & .60 & .25 & .67 & .14 & .23 & .15 & .74 & .25 \\
victim      & .70 & .88 & \textbf{.78} & .58 & .43 & .49 & .58 & .67 & .62 & .77 & .48 & .59 & .00 & .00 & .00 & .13 & .60 & .22 & .13 & .74 & .22 & .25 & .93 & \textbf{.39} \\
expert      & .93 & .62 & \textbf{.74} & .48 & .50 & .49 & .51 & .64 & .57 & .79 & .62 & .69 & .20 & .40 & .27 & .26 & .12 & .16 & .57 & .67 & \textbf{.62} & .19 & .98 & .32 \\
rules       & .82 & .43 & .56 & .49 & 1.00 & .66 & .86 & .57 & \textbf{.69} & .58 & .81 & .67 & .15 & .33 & .20 & .00 & .00 & .00 & .38 & .37 & \textbf{.38} & .13 & .86 & .23 \\
\bottomrule
\end{tabular}
\end{table*}

\begin{table*}[!t]
\caption{Zero-shot results of the fine-tuned models, $\text{config}_b$ (60/10/30). Best F1 per setting in bold.}
\label{tab:zeroshot-configB}
\centering
\small
\setlength{\tabcolsep}{2pt}
\renewcommand{\arraystretch}{1.05}
\begin{tabular}{@{}l*{12}{c}@{\hspace{4pt}}*{12}{c}@{}}
\toprule
& \multicolumn{12}{c}{\textbf{Balanced}}
& \multicolumn{12}{c}{\textbf{Imbalanced}} \\
\cmidrule(lr){2-13} \cmidrule(lr){14-25}
& \multicolumn{3}{c}{\textbf{LLaMA}} & \multicolumn{3}{c}{\textbf{Mistral}} & \multicolumn{3}{c}{\textbf{Gemma}} & \multicolumn{3}{c}{\textbf{Qwen}} & \multicolumn{3}{c}{\textbf{LLaMA}} & \multicolumn{3}{c}{\textbf{Mistral}} & \multicolumn{3}{c}{\textbf{Gemma}} & \multicolumn{3}{c}{\textbf{Qwen}} \\
\cmidrule(lr){2-4} \cmidrule(lr){5-7} \cmidrule(lr){8-10} \cmidrule(lr){11-13} \cmidrule(lr){14-16} \cmidrule(lr){17-19} \cmidrule(lr){20-22} \cmidrule(lr){23-25}
\textbf{Pattern} & \textbf{P} & \textbf{R} & \textbf{F1} & \textbf{P} & \textbf{R} & \textbf{F1} & \textbf{P} & \textbf{R} & \textbf{F1} & \textbf{P} & \textbf{R} & \textbf{F1} & \textbf{P} & \textbf{R} & \textbf{F1} & \textbf{P} & \textbf{R} & \textbf{F1} & \textbf{P} & \textbf{R} & \textbf{F1} & \textbf{P} & \textbf{R} & \textbf{F1} \\
\midrule
vanilla     & .50 & .99 & \textbf{.67} & .48 & .38 & .42 & .52 & .50 & .51 & .50 & 1.00 & \textbf{.67} & .20 & .02 & .04 & .50 & .01 & .02 & .79 & .21 & \textbf{.33} & .13 & .73 & .22 \\
choice      & .50 & .99 & .66 & .48 & .73 & .58 & .49 & .46 & .48 & .50 & .99 & \textbf{.67} & .15 & .66 & \textbf{.24} & .09 & .07 & .08 & .00 & .00 & .00 & .21 & .15 & .17 \\
cloze       & .50 & .96 & \textbf{.66} & .51 & .72 & .60 & .70 & .33 & .45 & .52 & .91 & \textbf{.66} & .12 & .73 & .21 & .24 & .46 & \textbf{.31} & .70 & .05 & .10 & .11 & .15 & .13 \\
cot         & .55 & .77 & .64 & .33 & .02 & .03 & .65 & .50 & .57 & .50 & .98 & \textbf{.66} & .00 & .00 & .00 & .30 & .41 & \textbf{.35} & .18 & .25 & .21 & .80 & .12 & .22 \\
target      & .52 & .96 & .67 & .54 & .65 & .59 & .47 & .50 & .49 & .53 & .95 & \textbf{.68} & .15 & .19 & \textbf{.17} & .89 & .06 & .12 & .50 & .01 & .02 & .60 & .02 & .05 \\
illocution.  & .55 & .97 & .70 & .50 & .71 & .59 & .61 & .96 & \textbf{.74} & .56 & .97 & .71 & .35 & .38 & \textbf{.36} & .13 & .20 & .16 & .28 & .31 & .30 & .19 & .73 & .30 \\
functional  & .75 & .82 & \textbf{.78} & .53 & .86 & .65 & .54 & .54 & .54 & .54 & .80 & .65 & .26 & .38 & .31 & .48 & .09 & .16 & .75 & .45 & \textbf{.56} & .29 & .46 & .35 \\
definition  & .53 & .94 & \textbf{.68} & .49 & .62 & .55 & .51 & .61 & .55 & .66 & .57 & .61 & .17 & .01 & .01 & .14 & .45 & .21 & .00 & .00 & .00 & .18 & .65 & \textbf{.28} \\
victim      & .51 & .39 & .44 & .51 & .79 & \textbf{.62} & .53 & .58 & .56 & .50 & .79 & .61 & .33 & .01 & .02 & .26 & .36 & .30 & .27 & .89 & \textbf{.41} & .21 & .38 & .27 \\
expert      & .74 & .93 & \textbf{.82} & .49 & .63 & .55 & .51 & .43 & .47 & .62 & .91 & .74 & .67 & .26 & .37 & .15 & .50 & .23 & .00 & .00 & .00 & .42 & .36 & \textbf{.39} \\
rules       & .55 & .66 & .60 & .54 & .70 & .61 & .49 & .51 & .50 & .51 & .99 & \textbf{.67} & .23 & .16 & .19 & .22 & .33 & .26 & .40 & .25 & \textbf{.31} & .13 & .78 & .22 \\
\bottomrule
\end{tabular}
\end{table*}

\begin{table*}[!t]
\caption{8-shot results of the fine-tuned models, balanced evaluation. Best F1 per configuration in bold.}
\label{tab:fewshot-finetuned}
\centering
\small
\setlength{\tabcolsep}{2pt}
\renewcommand{\arraystretch}{1.05}
\begin{tabular}{@{}l*{12}{c}@{\hspace{4pt}}*{12}{c}@{}}
\toprule
& \multicolumn{12}{c}{\textbf{Configuration~A (80/10/10)}}
& \multicolumn{12}{c}{\textbf{Configuration~B (60/10/30)}} \\
\cmidrule(lr){2-13} \cmidrule(lr){14-25}
& \multicolumn{3}{c}{\textbf{LLaMA}}
& \multicolumn{3}{c}{\textbf{Mistral}}
& \multicolumn{3}{c}{\textbf{Gemma}}
& \multicolumn{3}{c}{\textbf{Qwen}}
& \multicolumn{3}{c}{\textbf{LLaMA}}
& \multicolumn{3}{c}{\textbf{Mistral}}
& \multicolumn{3}{c}{\textbf{Gemma}}
& \multicolumn{3}{c}{\textbf{Qwen}} \\
\cmidrule(lr){2-4} \cmidrule(lr){5-7} \cmidrule(lr){8-10} \cmidrule(lr){11-13}
\cmidrule(lr){14-16} \cmidrule(lr){17-19} \cmidrule(lr){20-22} \cmidrule(lr){23-25}
\textbf{Pattern}
& \textbf{P} & \textbf{R} & \textbf{F1}
& \textbf{P} & \textbf{R} & \textbf{F1}
& \textbf{P} & \textbf{R} & \textbf{F1}
& \textbf{P} & \textbf{R} & \textbf{F1}
& \textbf{P} & \textbf{R} & \textbf{F1}
& \textbf{P} & \textbf{R} & \textbf{F1}
& \textbf{P} & \textbf{R} & \textbf{F1}
& \textbf{P} & \textbf{R} & \textbf{F1} \\
\midrule
vanilla      & .89 & .57 & \textbf{.70} & .64 & .33 & .44 & .60 & .62 & .61 & .52 & .76 & .62 & .73 & .82 & \textbf{.77} & .55 & .76 & .63 & .73 & .54 & .62 & .56 & .95 & .71 \\
choice       & .51 & .95 & .66 & .38 & .36 & .37 & .79 & .81 & \textbf{.80} & .54 & .86 & .66 & .50 & 1.00 & .66 & .50 & .85 & .63 & .66 & .98 & \textbf{.79} & .53 & .98 & .68 \\
cloze        & .50 & .98 & \textbf{.66} & .94 & .40 & .57 & .60 & .50 & .55 & .47 & .88 & .62 & .51 & .99 & .67 & .50 & .87 & .63 & .78 & .90 & \textbf{.84} & .58 & .76 & .66 \\
cot          & .92 & .57 & .71 & .65 & .26 & .37 & 1.00 & .29 & .44 & .73 & .86 & \textbf{.79} & .50 & 1.00 & .66 & .00 & .00 & .00 & .62 & .87 & \textbf{.73} & .53 & .95 & .68 \\
target       & .89 & .74 & \textbf{.81} & .63 & .52 & .57 & .36 & .33 & .35 & .83 & .71 & .77 & .61 & .91 & \textbf{.73} & .52 & .97 & .68 & .49 & .41 & .44 & .52 & .76 & .62 \\
illocution.  & .57 & .60 & .58 & .61 & .33 & .43 & .70 & .93 & \textbf{.80} & .90 & .62 & .73 & .50 & .81 & .62 & .49 & .61 & .54 & .57 & .46 & .51 & .51 & .91 & \textbf{.65} \\
functional   & .69 & .90 & .78 & .53 & .57 & .55 & .91 & .76 & \textbf{.83} & .86 & .76 & .81 & .54 & .95 & .69 & .50 & 1.00 & .67 & .53 & .51 & .52 & .61 & .94 & \textbf{.74} \\
definition   & .79 & .36 & .49 & .64 & .17 & .26 & .59 & .62 & .60 & .80 & .79 & \textbf{.80} & .52 & .61 & .56 & .50 & 1.00 & \textbf{.66} & .68 & .37 & .48 & .50 & .80 & .61 \\
victim       & .88 & .55 & \textbf{.68} & .50 & 1.00 & .67 & .56 & .69 & .62 & .51 & .88 & .65 & .77 & .70 & \textbf{.73} & .50 & 1.00 & .66 & .49 & .83 & .62 & .64 & .80 & .71 \\
expert       & .58 & .88 & .70 & .50 & .69 & .58 & .79 & .71 & \textbf{.75} & .65 & .71 & .68 & .82 & .43 & .57 & .49 & .84 & .62 & .73 & .63 & .68 & .59 & .87 & \textbf{.71} \\
rules        & .93 & .62 & .74 & .53 & .24 & .33 & .91 & .76 & \textbf{.83} & .78 & .74 & .76 & .47 & .73 & .57 & .53 & .74 & \textbf{.62} & .50 & .50 & .50 & .49 & .78 & .60 \\
\bottomrule
\end{tabular}
\end{table*}

Finally, we evaluate the Balanced-Train / Imbalanced-Test setting, in which models are trained on the balanced set but tested on the imbalanced set, with no overlap between the two. Results are reported in Table~\ref{tab:cross-balanced}, and the BERT classification heads trained on the same balanced data reach at most 0.26 F1 under $\text{config}_a$ and 0.27 F1 under $\text{config}_b$ on the same test sets (table in the repository).
As expected, balanced training substantially improves performance on the imbalanced test set in 61\% of cases, a setting more closely aligned with realistic deployment scenarios for Danish offensive language detection.
Notably, in the best case, Gemma with the illocutionary prompt achieves 0.70 F1 on the imbalanced test set despite its heavily skewed class distribution (13\% offensive).
We further observe that 68\% of
$\text{config}_a$ cases outperform their $\text{config}_b$ counterparts, suggesting a positive relationship between training set size and performance.

\begin{table*}[!t]
\caption{Balanced-Train / Imbalanced-Test zero-shot results. Best F1 per configuration in bold.}
\label{tab:cross-balanced}
\centering
\small
\setlength{\tabcolsep}{2pt}
\renewcommand{\arraystretch}{1.05}
\begin{tabular}{@{}l*{12}{c}@{\hspace{4pt}}*{12}{c}@{}}
\toprule
& \multicolumn{12}{c}{\textbf{Configuration~A (80/10/10)}}
& \multicolumn{12}{c}{\textbf{Configuration~B (60/10/30)}} \\
\cmidrule(lr){2-13} \cmidrule(lr){14-25}
& \multicolumn{3}{c}{\textbf{LLaMA}}
& \multicolumn{3}{c}{\textbf{Mistral}}
& \multicolumn{3}{c}{\textbf{Gemma}}
& \multicolumn{3}{c}{\textbf{Qwen}}
& \multicolumn{3}{c}{\textbf{LLaMA}}
& \multicolumn{3}{c}{\textbf{Mistral}}
& \multicolumn{3}{c}{\textbf{Gemma}}
& \multicolumn{3}{c}{\textbf{Qwen}} \\
\cmidrule(lr){2-4} \cmidrule(lr){5-7} \cmidrule(lr){8-10} \cmidrule(lr){11-13}
\cmidrule(lr){14-16} \cmidrule(lr){17-19} \cmidrule(lr){20-22} \cmidrule(lr){23-25}
\textbf{Pattern}
& \textbf{P} & \textbf{R} & \textbf{F1}
& \textbf{P} & \textbf{R} & \textbf{F1}
& \textbf{P} & \textbf{R} & \textbf{F1}
& \textbf{P} & \textbf{R} & \textbf{F1}
& \textbf{P} & \textbf{R} & \textbf{F1}
& \textbf{P} & \textbf{R} & \textbf{F1}
& \textbf{P} & \textbf{R} & \textbf{F1}
& \textbf{P} & \textbf{R} & \textbf{F1} \\
\midrule
vanilla      & .25 & .70 & .37 & .20 & .42 & .27 & .16 & .86 & .28 & .34 & .70 & \textbf{.45} & .15 & .17 & .16 & .14 & .70 & \textbf{.24} & .14 & .41 & .21 & .17 & .11 & .13 \\
choice       & .32 & .84 & \textbf{.46} & .14 & .79 & .24 & .22 & .63 & .33 & .17 & .91 & .29 & .13 & .96 & .24 & .13 & .49 & .20 & .24 & .38 & \textbf{.29} & .13 & .98 & .23 \\
cloze        & .18 & .91 & .30 & .22 & .86 & .34 & .85 & .26 & \textbf{.39} & .13 & 1.00 & .24 & .15 & .82 & .25 & .14 & .59 & .23 & .21 & .71 & \textbf{.32} & .14 & .81 & .24 \\
cot          & .21 & .30 & .25 & .13 & 1.00 & .23 & .28 & .60 & .39 & .31 & .81 & \textbf{.45} & .20 & .45 & .27 & .13 & .36 & .19 & .29 & .34 & \textbf{.32} & .17 & .80 & .29 \\
target       & .27 & .93 & \textbf{.42} & .13 & 1.00 & .23 & .13 & 1.00 & .23 & .15 & .91 & .26 & .34 & .09 & .14 & .13 & .15 & .14 & .35 & .66 & \textbf{.46} & .20 & .65 & .31 \\
illocution.  & .22 & .95 & .35 & .12 & .65 & .21 & .67 & .72 & \textbf{.70} & .16 & .81 & .27 & .15 & .24 & .18 & .14 & .34 & .20 & .51 & .55 & \textbf{.53} & .21 & .35 & .26 \\
functional   & .27 & .91 & .41 & .16 & .19 & .17 & .34 & .88 & \textbf{.49} & .19 & .91 & .31 & .47 & .55 & .51 & .21 & .62 & .32 & .44 & .84 & \textbf{.58} & .16 & .90 & .27 \\
definition   & .20 & .86 & .33 & .19 & .70 & .30 & .24 & .93 & \textbf{.39} & .17 & .33 & .22 & .39 & .12 & .19 & .12 & .29 & .17 & .14 & .52 & .22 & .18 & .82 & \textbf{.30} \\
victim       & .25 & .84 & .39 & .11 & .51 & .18 & .33 & .86 & \textbf{.48} & .17 & .95 & .29 & .49 & .45 & \textbf{.47} & .16 & .78 & .26 & .13 & .46 & .20 & .40 & .46 & .43 \\
expert       & .23 & .81 & .36 & .13 & .37 & .19 & .48 & .81 & \textbf{.60} & .00 & .00 & .00 & .29 & .73 & .41 & .13 & 1.00 & .23 & .50 & .47 & \textbf{.48} & .31 & .20 & .24 \\
rules        & .22 & .65 & .33 & .18 & .79 & .29 & .48 & .77 & \textbf{.59} & .18 & 1.00 & .30 & .21 & .38 & \textbf{.27} & .14 & .29 & .19 & .12 & .46 & .19 & .16 & .05 & .08 \\
\bottomrule
\end{tabular}
\end{table*}

\subsection{Cross-Prompt Generalization}

\begin{figure}[!t]
    \centering
    \subfloat[Cross-prompt transfer for LLaMA\label{subfig:heatmap-llama}]{%
        \includegraphics[width=\columnwidth]{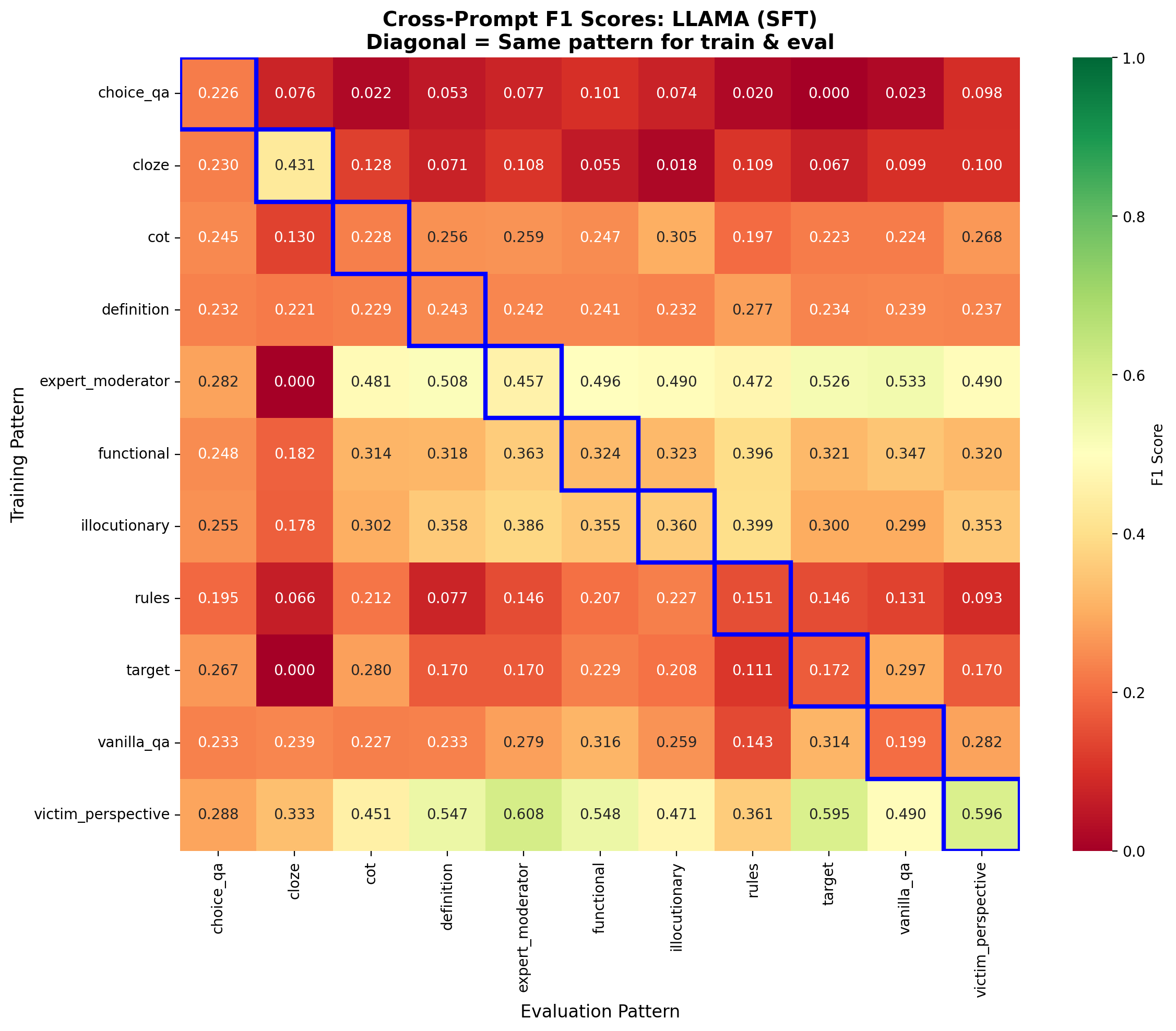}}\\
    \subfloat[Cross-prompt transfer for Gemma\label{subfig:heatmap-gemma}]{%
        \includegraphics[width=\columnwidth]{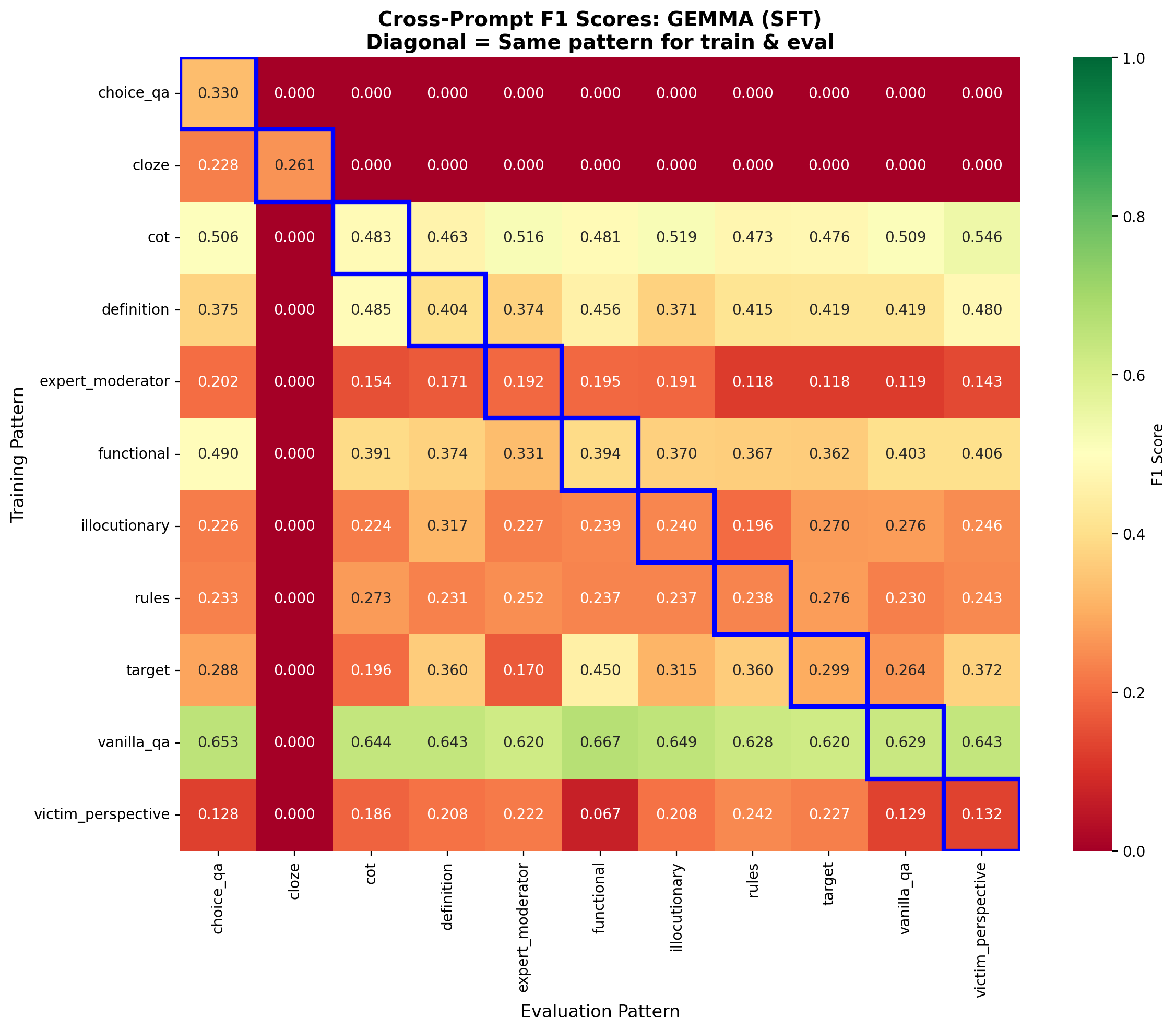}}
    \caption{Cross-prompt transfer matrices for LLaMA and Gemma, Imbalanced setting.}
    \label{fig:cross-prompt-imbalanced-1}
\end{figure}

Having established that prompt selection significantly affects within-pattern performance, we now analyze how the choice of fine-tuning prompt transfers across prompting patterns.
Specifically, we construct this evaluation by fine-tuning a model variant on each of the eleven patterns and testing it against all remaining ten patterns.
Figure~\ref{fig:cross-prompt-imbalanced-1} shows the imbalanced-setting matrices for LLaMA and Gemma, and the matrices for Mistral and Qwen, together with all four balanced-setting matrices, are in the repository.
We have two consistent findings.
First, format-constrained patterns (choice QA, cloze) exhibit near-zero cross-pattern transfer. Intuitively, this is as expected because each of these patterns enforces a specific response format that is not compatible with other formats.
Moreover, besides guiding the model to respond in Danish, this type of prompt also forces the LLM to adopt a new response format, which may conflict with what the model has learned from extensive SFT and RLHF on higher-resource languages.
Second, in contrast, semantically grounded patterns that frame classification in terms of intent, impact, or perspective exhibit more robust cross-pattern transfer, though the best-performing source pattern varies by model.

\subsection{Danish Orthographic Effects}\label{subsec:rq3}

We next examine whether replacing Danish-specific characters (\texttt{æ}, \texttt{ø}, \texttt{å}) with their ASCII transliterations (\texttt{ae}, \texttt{oe}, \texttt{aa}) affects model performance, as described in Section~\ref{subsec:dataset}. Table~\ref{tab:robustness} (top) reports the results. These runs, like the denoising and cross-prompt experiments, were fine-tuned with a 4,096-token maximum sequence length rather than the 512 tokens of the main tables, and the original-orthography and standard-prompt rows were decoded with a 100-token output budget against 20 tokens for the ASCII and denoising rows, so the two arms of each comparison differ in decoding budget as well as in input, a confound we flag rather than remove.
Contrary to the expectation that ASCII normalization might help models with limited Danish exposure, transliteration generally degrades performance on imbalanced data, with Gemma exhibiting the largest drop (0.486 to 0.258 mean F1).
This suggests that multilingual pre-training has established meaningful representations for these characters, and removing them discards useful signal.
Qwen is a notable exception, showing slight improvement with ASCII transliteration on imbalanced data. These findings align with Müller-Eberstein et al. \cite{muellereberstein2025dakultur}, who show that native Danish data outperforms translated alternatives.  To examine this orthographic effect at the subword level, we measure how each model's tokenizer fragments Danish words. Table~\ref{tab:tokenization} reports, per model, the average number of subword tokens for words containing \texttt{æ}, \texttt{ø}, or \texttt{å} versus words without them. Across all four models, Danish-specific characters trigger substantially more fragmentation (a ratio of 1.40--1.68$\times$), with Mistral's smaller 32K vocabulary most affected and Gemma's 256K vocabulary least. This shows that ASCII transliteration alters the input at the tokenization level rather than preserving it, offering a mechanism consistent with the performance degradation above. We do not claim a direct causal link between tokenization efficiency and F1, as our four-model sample is too small to support such a claim and Qwen is a partial counterexample.

\begin{table}[!t]
\caption{Average subword tokens per Danish word with and without \texttt{æ}/\texttt{ø}/\texttt{å}, and their ratio, per tokenizer.}
\label{tab:tokenization}
\centering
\small
\setlength{\tabcolsep}{4pt}
\renewcommand{\arraystretch}{1.05}
\begin{tabular}{@{}lrrrr@{}}
\toprule
\textbf{Model} & \textbf{Vocab} & \textbf{w/ æøå} & \textbf{w/o æøå} & \textbf{Ratio} \\
\midrule
LLaMA   & 128K & 2.64 & 1.68 & 1.57 \\
Mistral & 32K  & 3.18 & 1.89 & 1.68 \\
Gemma   & 256K & 2.09 & 1.50 & 1.40 \\
Qwen    & 152K & 2.65 & 1.70 & 1.56 \\
\bottomrule
\end{tabular}
\end{table}

\begin{table}[!t]
\caption{Original orthography vs.\ ASCII transliteration (top) and standard vs.\ denoising prompts (bottom). Best value per column in bold.}
\label{tab:robustness}
\centering
\small
\renewcommand{\arraystretch}{1.05}
\begin{tabular*}{\columnwidth}{@{\extracolsep{\fill}} lcccc @{}}
\toprule
& \textbf{LLaMA} & \textbf{Mistral} & \textbf{Gemma} & \textbf{Qwen} \\
\midrule
\multicolumn{5}{l}{\textit{Orthographic variation (F1 averaged across patterns)}} \\
\midrule
\multicolumn{5}{c}{\textit{Balanced}} \\
Original (æøå)   & .594 & .501 & .679 & \textbf{.691} \\
ASCII (aeoeaa)   & \textbf{.597} & \textbf{.562} & .550 & .579 \\
\multicolumn{5}{c}{\textit{Imbalanced}} \\
Original (æøå)   & \textbf{.267} & \textbf{.283} & \textbf{.486} & .227 \\
ASCII (aeoeaa)   & .187 & .149 & .258 & \textbf{.250} \\
\midrule
\multicolumn{5}{l}{\textit{Denoising prompts (F1 averaged across patterns)}} \\
\midrule
\multicolumn{5}{c}{\textit{Balanced}} \\
Standard         & \textbf{.594} & .501 & \textbf{.679} & \textbf{.691} \\
Denoising        & .547 & \textbf{.555} & .600 & .582 \\
\multicolumn{5}{c}{\textit{Imbalanced}} \\
Standard         & \textbf{.267} & \textbf{.283} & \textbf{.486} & .227 \\
Denoising        & .220 & .155 & .263 & \textbf{.261} \\
\bottomrule
\end{tabular*}
\end{table}

\subsection{Preprocessing Strategies}
Finally, we evaluate whether explicitly instructing models to ignore social media noise (e.g., misspellings, slang, non-standard punctuation) improves classification performance.
Table~\ref{tab:robustness} (bottom) compares standard prompts against denoising variants.
The results show that denoising actually degrades performance in 75\% (6 of 8) cases.
Furthermore, the harm can be substantial, e.g., a 45.8\% performance drop for Gemma in the imbalanced setting.
These results suggest that instruction-tuned LLMs already handle social media noise effectively, and adding explicit denoising instructions diverts attention away from the classification objective.

\section{Discussion}
Two patterns in our results carry practical weight. First, balanced and imbalanced evaluation tell different stories: the best fine-tuned F1 falls from 0.84 under balanced evaluation ($\text{config}_a$, Table~\ref{tab:zeroshot-configA}) to 0.62 under the corpus's natural 13\% class skew -- a 26.2\% relative drop -- and no model--prompt combination is uniformly robust across settings. Balanced-only evaluation therefore risks overstating deployment readiness, and we recommend that results for low-resource offensive language detection always be reported under natural class distributions as well. Second, model--prompt interactions dominate: the best pattern differs by model and by setting, format-constrained patterns fail to transfer across prompts (Figure~\ref{fig:cross-prompt-imbalanced-1}), and theory-informed prompts are frequently -- though not uniformly -- the more robust choice under skew. For practitioners, this argues for validating a small pool of semantically grounded prompts per model rather than adopting a single universal template.

The BERT results should be read with their scale and training regime in mind. Our baselines are 110M--124M-parameter encoders fine-tuned with lightweight LoRA adapters; DaBERT's advantage over multilingual alternatives on imbalanced data suggests monolingual pre-training retains value, while ScandiBERT's balanced-data score reflects a degenerate all-positive classifier rather than learned discrimination. Stronger encoder training regimes could plausibly narrow the gap to LLMs, and we regard the LLM-versus-encoder comparison at matched tuning effort as an open question that our results motivate rather than settle.

\section{Conclusion}

We presented a systematic evaluation of open-weight LLMs and BERT models for Danish offensive language detection, spanning four LLM families, three BERT baselines, eleven prompting patterns across five cognitive framing categories, three evaluation settings, and two split configurations.
Our experiments reveal that general-purpose LLMs surpass our fine-tuned BERT baselines even without task-specific fine-tuning, provided a suitable prompting pattern is used. We caution, however, that the LLMs and BERT baselines differ in parameter count by nearly two orders of magnitude, so this comparison reflects the joint effect of pre-training breadth and model capacity rather than pre-training strategy alone.
Lightweight fine-tuning yields substantial further gains, whereas few-shot prompting improves performance less often after fine-tuning than before it (55\% vs.\ 65\% of configurations), consistent with diminishing returns once task-specific adaptation has already occurred.

On the prompting side, theory-driven and perspective-based patterns are often comparatively robust under class imbalance, though no prompt category dominates uniformly, and individual LLM families exhibit distinct preferences for specific prompting categories, indicating that effective prompt selection is model-dependent rather than universal.
Cross-prompt transfer analysis further reveals that format-constrained patterns fail to generalize across prompts, while semantically grounded patterns that frame classification in terms of intent, impact, or perspective transfer more robustly.
Both ASCII transliteration of Danish-specific characters (\texttt{æ}, \texttt{ø}, \texttt{å}) and explicit denoising instructions degrade performance in most settings, indicating that these models have already internalized meaningful representations for Danish orthography and social media noise during pre-training.
Future work should explore the construction of larger Danish offensive language corpora, as our results suggest a positive relationship between training set size and detection performance, as well as multi-class annotation distinguishing fine-grained categories such as racism, sexism, and religious hatred.

\section{Limitations}
As the only publicly available annotated dataset for Danish offensive language, the DKHate corpus contains approximately 3,300 samples, of which only 425 are offensive.
While this is sufficient to yield meaningful findings, as demonstrated by the consistent patterns observed across our experiments, a larger corpus with more offensive samples would enable more robust evaluation, particularly under imbalanced conditions.
We evaluate only four open-weight LLMs in the 7–9B parameter range.
In future work, we plan to extend this evaluation to larger models within these families.
Additionally, the corpus provides only binary labels, limiting analysis to coarse-grained classification.
Our deterministic output parser maps non-conforming model outputs to the negative class rather than categorizing them separately; this means we cannot distinguish format non-compliance from genuine classification errors in all cases, and on class-skewed data degenerate outputs can appear deceptively accurate. A post-hoc audit of saved outputs (Table~\ref{tab:parse-failure}) further indicates that this parser is unreliable for the choice QA pattern specifically: its substring-matching rule mislabels roughly 9\% of otherwise well-formed choice QA responses, while every other pattern is affected at a negligible rate. We therefore treat choice QA results as indicative and base our conclusions on the remaining ten patterns.
Three further limitations concern measurement and scope. First, all reported values are single-run point estimates obtained from a single run on fixed splits, with sampling-based decoding and no fixed generation seed; we report no confidence intervals or significance tests, and on test sets as small as 85 examples (balanced $\text{config}_a$) differences of a few points of F1 are within sampling noise and should not be over-interpreted. Second, because we report positive-class F1, our figures are not directly comparable to the macro-averaged F1 conventionally reported in prior work on this corpus. The published macro-F1 of 0.699 for the best DKHate system, a logistic regression, rests on an F1 of 0.921 for the not-offensive class and of 0.476 for the offensive class \cite{sigurbergsson-derczynski-2020-offensive}, so the comparable published positive-class figure is 0.48, against our best imbalanced results of 0.62 after fine-tuning and 0.45 zero-shot (Tables~\ref{tab:zeroshot-configA} and~\ref{tab:zeroshot-base-A}), obtained on a different split of a smaller release of the corpus. Third, our results are a snapshot of a moving landscape, since the four checkpoints are fixed releases from 2024, the runs were made between December 2025 and March 2026, and the DKHate comments were collected by 2019, while both the models and the vocabulary of Danish offensive language keep changing as new terms and targets enter with current events. The comparative patterns we report are therefore scoped to these releases and to this collection period, and a successor model or a freshly collected corpus could move the absolute numbers even where the pattern-level findings hold.

\section*{Reproducibility}
All code, configurations, datasets and instructions for reproducing the experiments reported in this paper, including BERT model evaluation and LLM fine-tuning pipelines, are publicly available at \url{https://github.com/PASHAADC/Danish-LLM-HateSpeech}.

\section*{Ethical Considerations}
A further consideration concerns the dual-use nature of this work. The open-weight models and lightweight fine-tuning procedure we study are intended to support detection and moderation of offensive language, but the same techniques could in principle be repurposed to generate such content or to help adversaries craft text that evades detection. We view the benefits of open, reproducible research on low-resource offensive language detection---particularly for an underserved language such as Danish---as outweighing this risk, especially as the underlying models and methods are already publicly available. To discourage misuse, we release our code and configurations under a license that restricts use to research and content-moderation purposes and explicitly prohibits applications intended to generate offensive content or circumvent its detection.

\bibliographystyle{IEEEtran}
\bibliography{references}

\end{document}